\documentclass[a4paper,conference]{IEEEtran}

\usepackage{cite}
\usepackage[pdftex]{graphicx}
\usepackage{amsmath}
\usepackage{amsfonts}
\usepackage{array}
\usepackage[ruled,vlined]{algorithm2e}
\usepackage[caption=false,font=normalsize,labelfont=sf,textfont=sf]{subfig}
\usepackage{booktabs} 
\usepackage{tabularx}
\usepackage{multirow} 
\usepackage{hyperref}
\usepackage{url}
\usepackage{siunitx}
\begin{document}
    
    \title{Understanding Integrated Gradients with SmoothTaylor for Deep Neural Network Attribution}
    
    \author{\IEEEauthorblockN{Gary S. W. Goh\IEEEauthorrefmark{2}, Sebastian Lapuschkin\IEEEauthorrefmark{3}, Leander Weber\IEEEauthorrefmark{3}, Wojciech Samek\IEEEauthorrefmark{3} and Alexander Binder\IEEEauthorrefmark{2}}
        \IEEEauthorblockA{\IEEEauthorrefmark{2}ISTD Pillar, Singapore University of Technology and Design, Singapore 487372}
        \IEEEauthorblockA{\IEEEauthorrefmark{3}Fraunhofer Heinrich Hertz Institute, 10587 Berlin, Germany}
        \texttt{gary\_goh@mymail.sutd.edu.sg, alexander\_binder@sutd.edu.sg}\\
        \texttt{\string{sebastian.lapuschkin, leander.weber, wojciech.samek\string}@hhi.fraunhofer.de}
    }
    
    \maketitle
    
    \begin{abstract}
        \textit{Integrated Gradients} as an attribution method for deep neural network models offers simple implementability.
        However, it suffers from noisiness of explanations which affects the ease of interpretability.
        The \textit{SmoothGrad} technique is proposed to solve the noisiness issue and smoothen the attribution maps of any gradient-based attribution method.
        In this paper, we present \textit{SmoothTaylor} as a novel theoretical concept bridging \textit{Integrated Gradients} and \textit{SmoothGrad}, from the Taylor's theorem perspective.
        We apply the methods to the image classification problem, using the ILSVRC2012 ImageNet object recognition dataset, and a couple of pretrained image models to generate attribution maps.
        These attribution maps are empirically evaluated using quantitative measures for sensitivity and noise level.
        We further propose adaptive noising to optimize for the noise scale hyperparameter value.
        From our experiments, we find that the \textit{SmoothTaylor} approach together with adaptive noising is able to generate better quality saliency maps with lesser noise and higher sensitivity to the relevant points in the input space as compared to \textit{Integrated Gradients}.
    \end{abstract}
    
    \section{Introduction}
    Deep neural networks have displayed remarkable success in various large-scale, real-world and complex artificial intelligence tasks in computer vision \cite{Huang2017,Ren2017,Tan2019} and natural language processing \cite{Edunov2018, Johnson2017}.
    However, these high performing non-linear neural models, unlike traditional machine learning models, act like a \textit{black box} which suffers from poor input-to-output inference and interpretability.
    Due to the nature of how deep neural network algorithms are designed, it is difficult to explain \textit{what} or \text{why} an \textit{individual} input result in the model arriving at a particular output \cite{Fan2020}.
    This major disadvantage hinders human experts to fully understand the basis and the reasoning of every prediction a deep neural model makes for each input, limiting the extent of its application in practice.
    
    With the aim to better understand the complex input-to-output behavior of a deep neural network, a number of previous work \cite{Baehrens2010,Simonyan2014,Zeiler2014,Springenberg2015,Zhou2016,Zintgraf2016,Binder2016,Shrikumar2017,Sundararajan2017,Montavon2017,Selvaraju2020,SamXAI19} focus on the problem of attribution.
    Attributions measure the contribution of the model's output explained in terms of its input variables.
    For instance, for image classification systems, an attribution method assigns a relevance score to every pixel of the input image that explains for the model's predicted class.
    There are many applications where such an ability to ``explain" for a complex model's decision is crucial.
    Attributions act as supporting evidence to explain the rationale of a model's decision.
    This helps to facilitate the building of trust between humans and automated systems \cite{Gilpin2019}, and encourage higher adoption of deep neural networks in practice, especially in high-risk application areas.
    The importance of attribution is more apparent in view of the recent vulnerability discoveries in deep neural networks against malicious and yet unnoticeable to-the-human-eye adversarial attacks \cite{Nguyen2015,Moosavi-Dezfooli2017}.
    
    Sundararajan et al. \cite{Sundararajan2017} proposed \textit{Integrated Gradients} (\textit{IG}) as an attribution method for deep neural networks, which unlike other methods \cite{Zeiler2014,Springenberg2015,Binder2016,Zhou2016,Shrikumar2017,Montavon2017,Selvaraju2020}, is fully independent of the composition of the model's structure, and can be easily implemented with access to just the input's gradients after back-propagation.
    As such, it is computationally efficient to compute, and can be widely applied to various deep neural networks architectures and tasks.
    
\begin{figure*}[!t]
        \centering
        \includegraphics[width=0.95\linewidth]{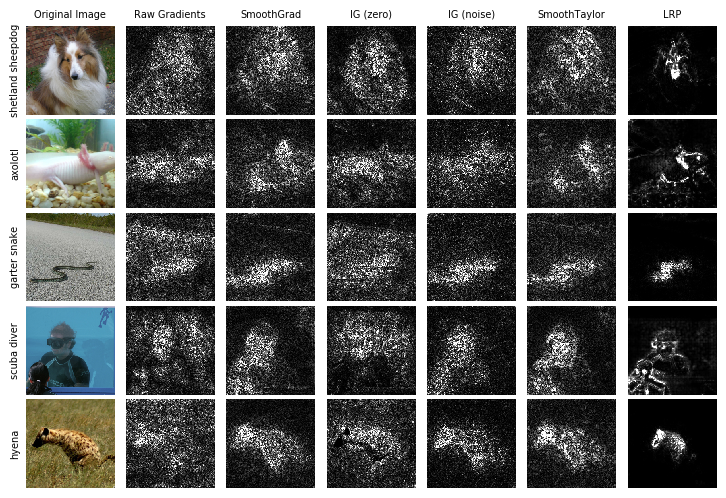}
        \caption{Comparison of saliency maps computed by different attribution methods. These saliency maps show the relative contributions of each input pixel that explains for the model's prediction. \underline{Columns from the left}: original input image; raw gradients; \textit{SmoothGrad}; \textit{IG} with zero as the baseline ($M=50$); \textit{IG} with noise as the baseline ($N=1$); \textit{SmoothTaylor} ($\sigma=$\num{5e-1}, $R=150$); \textit{Layer-wise Relevance Propagation}. \underline{Setup}: DenseNet121 image classifier pretrained for ImageNet. Normalized absolute values are used to visualize the attribution maps and values above 99\textsuperscript{th} percentile are clipped.}
        \label{fig:method-comparison}
    \end{figure*}

    However, \textit{IG} require a selected baseline as a benchmark, which raises the question on how such a baseline is to be chosen.
    In addition, just as with other gradient-based methods \cite{Baehrens2010,Simonyan2014}, \textit{IG} often create attribution maps that are noisy which affects the ease of its interpretability.
    For example, compare the saliency maps (attribution maps visualized by a 2D image) of \textit{IG} (center two) with other methods \cite{Simonyan2014,Binder2016,Smilkov2017} in Figure \ref{fig:method-comparison}, which is based on a DenseNet \cite{Huang2017} with 121 layers pretrained for the ImageNet image classification task.
    The noisiness of its explanations is visually striking.
    
    Those noise pixels seemingly scattered at random across the maps as shown in Figure \ref{fig:method-comparison} may indeed reflect the true behavior of the gradients of the deep neural model: as the networks get deeper, the gradients across the input space fluctuate more sharply, resembling white noise, which is described as the shattering gradient problem \cite{Balduzzi2017}.
    To tackle the noisiness issue, Smilkov et al. \cite{Smilkov2017} proposed the \textit{SmoothGrad} technique, which uses a random sampling strategy around the input with averaging of the obtained attributions to produce visually sharper attribution maps.
    
    
    In this paper, our contributions are as follows:
    \begin{itemize}
        \item We present \textit{SmoothTaylor} as a theoretical concept bridge between \textit{IG} and \textit{SmoothGrad}.
        Unlike \textit{IG}, it does not require a selected fixed baseline.
        Under additional assumptions, \textit{SmoothTaylor} is an instance of \textit{SmoothGrad}.
        Regarding novelty, \textit{SmoothTaylor} is derived from the Taylor's theorem.
        Experimental results show that \textit{SmoothTaylor} is able to produce higher quality attribution maps that are more sensitive and less noisy as compared to \textit{IG}.
        \item From the perspective of gradient shattering, we explain why \textit{SmoothGrad} and \textit{SmoothTaylor} deteriorate with too small amount of added noise.
        \item We emphasize smoothness as a second quality measure for attribution and introduce  multi-scaled average total variation as a new evaluation measure for smoothness of the attribution maps.
        \item We further propose adaptive noising for individual input samples to optimize for either  predictor sensitivity of the generated attribution map or the noisiness of it. We show that it results in large improvements in performance compared to constant noise levels.
        \item This paper aims at a better understanding of existing gradient-based attribution methods.
    \end{itemize}
    
    The rest of the paper is organized as follows.
    Section \ref{sec:prelim} briefly describes \textit{IG} and \textit{SmoothGrad}.
    In Section \ref{sec:sig}, we derive \textit{SmoothTaylor} as a theoretical bridging concept.
    Next, in Section \ref{sec:experiment}, we conduct experiments by applying the attribution methods on a large-scale image classification problem to generate attribution maps.
    These attribution maps are quantitatively evaluated and compared.
    Adaptive noising is discussed in Section \ref{sec:adapt-noise}.
    
    \section{Preliminaries}\label{sec:prelim}
    
    \subsection{Integrated Gradients}
    
    Suppose one aims to explain the prediction of a deep neural network represented by a function $f$ for input $x$.
    The integrated gradient \cite{Sundararajan2017} for the $i^{th}$ dimension of the input is defined as follows:
    \begin{equation}
    IG_i(x, z) := (x_i - z_i) \times \int_{\alpha=0}^{1} \frac{\partial{f(z + \alpha \times (x - z))}}{\partial{x_i}} d\alpha
    \label{eqn:ig}
    \end{equation}
    
    The gradient of $f$ in the $i^{th}$ dimension is denoted by $\frac{\partial{f(x)}}{\partial{x_i}}$, and $z$ is a selected input baseline.
    In practice, the path integral is usually approximated by a summation across discrete small intervals $m$ with $M$ steps along the straightline path from input $x$ to baseline $z$, as follows:
    \begin{equation}
    IG_i(x, z) \approx (x_i - z_i) \times \frac{1}{M} \sum_{m=1}^{M} \frac{\partial{f(z + \frac{m}{M} \times (x - z))}}{\partial{x_i}}
    \label{eqn:ig-approx}
    \end{equation}
    
    Note that the attributions of the \textit{IG} method satisfy some desirable properties.
    First, it satisfies \textit{implementation invariance} since the computations are only based on the gradients of $f$, and are fully independent on any aspects of the models.
    It also fulfils the \textit{completeness} axiom, which ensures that the attributions add up to the output difference between input $x$ and baseline $z$ (i.e. $\sum_{i} IG_i(x, z) = f(x) - f(z)$).
    
    Thus, it is recommended to choose baseline $z$ to be zero (with a near-zero score, i.e. $f(z) \approx 0$) to represent the absence of input features.
    This acts as a basis for comparison and thus allows for the interpretation of the attributions to be a function of solely the individual input features.
    For images, this is a fully black image, which is argued to be a natural and intuitive choice.
    However, a black image is usually a statistical outlier to most pretrained models, which makes explanations relative to implausible outlier points seem irrelevant.
    Another disadvantage of using zero as the baseline is that input features that are zero or near-zero will never appear on the attribution maps since multiplier $x_i - z_i$ will be almost close to zero.
    For example in Figure \ref{fig:method-comparison}, saliency maps of \textit{IG} with zero as the baseline mostly fail to highlight objects of interests represented by dark-colored pixels.
    
    An alternative baseline with the same near-zero score property is also proposed -- uniform random noise.
    To address the issue of which random noise baseline to be chosen, a valid approach is to draw different noise baselines $z^{(n)}$ to compute $N$ \textit{IG} mappings, and average over them\footnote{\url{https://github.com/ankurtaly/Integrated-Gradients/}}:
    \begin{equation}
    \overline{IG}_{noise}(x) = \frac{1}{N} \sum_{n=1}^{N} IG(x, z^{(n)})
    \label{eqn:ig-noise}
    \end{equation}
    
    This slight extension does seem to improve \textit{IG} and result in more sensitive attribution maps with less noise, though there is still much room for improvement.
    Moreover, it should be noted that uniform random noise is also an unseen outlier, thus it guides to generate explanations that are no more meaningful than the zero baseline.
    Perhaps, the need for this method to fix a baseline that is consistent enough for all inputs, and at the same time does not deviate too far from the points in the dataset, is a fundamental flaw in its design, as such a baseline may not exist.
    
    \subsection{SmoothGrad}
    
    While the original \textit{SmoothGrad} technique \cite{Smilkov2017} smooths the raw gradients over the input space, it can be viewed as a general procedure which computes an attribution map by averaging over multiple attribution maps of an arbitrary gradient-based attribution method (denoted as $\mathcal{M}$) with multiple $N'$ noised inputs:
    \begin{equation}
    \label{eqn:smooth-grad}
    SmoothGrad(x) = \frac{1}{N'} \sum_{n=1}^{N'} \mathcal{M}(x + \epsilon),\ \epsilon \sim \mathcal{N}(0,\sigma'^2)
    \end{equation}
    
    Gaussian noise with parameter $\sigma'$ is used to smoothen the input space of the attribution method and construct visually sharper attribution maps.
    It is briefly discussed in their paper that $\sigma'$ needs to be carefully selected to get the best result.
    If too small, the attribution maps are still noisy; if too large, the maps become irrelevant.
    
    \section{SmoothTaylor}\label{sec:sig}
    
    In this section, we explain the derivation of \textit{SmoothTaylor}.
    Firstly, we discuss the motivation of our proposed improvement from the Taylor's theorem approximation perspective.
    Any arbitrary differentiable function $f$ can be approximated by Taylor's theorem with the first order term while ignoring all other higher order terms:
    \begin{equation}
    f(x) \approx f(z) + \sum_{i} (x_i - z_i)  \frac{\partial f(z)}{\partial x_i}
    \label{eqn:taylor-series}
    \end{equation}
    This yields an explanation, which describes how the output of the model $f(\cdot)$ in point $x$ is different from the output of the same model in point $z$.
    Notably, it is an explanation for $x$ relative to $z$. This raises the valid issue on how the point $z$ should be chosen.
    
    Secondly, in statistics, a valid method to deal with uncertainty is to compute an average over an uncertain quantity.
    In the case of uncertainty about which point $z$ should be chosen, the proper approach is to draw several roots $z^{(r)}$ (according to some method which we defer the discussion till later) and average over them, so as to improve the power of the approximation:
    \begin{equation}
    f(x) \approx \frac{1}{R} \sum_{r=1}^{R} \left[ f(z^{(r)}) + \sum_{i} (x_i - z^{(r)}_i)  \frac{\partial f(z^{(r)}))}{\partial x_i} \right]
    \label{eqn:taylor-series-average}
    \end{equation}
    Equation \eqref{eqn:taylor-series-average}, in turn, is a discrete approximation for the integral (with $S$ which has to be a measurable set):
    \begin{equation}
    f(x) \approx \int_{z \in S}  f(z) + \sum_{i}  (x_i - z_i) \frac{\partial f(z)}{\partial x_i}dz
    \label{eqn:taylor-series-average-integral}
    \end{equation}
    
    We are now ready to outline our method.
    Based on the concepts described above, the smooth integrated gradient in the $i^{th}$ dimension of an input $x$ within a set of roots $z \in S$ is defined as follows:
    \begin{equation}
    SmoothTaylor_i(x) := \int_{z \in S}  (x_i - z_i) \frac{\partial f(z)}{\partial x_i} dz
    \label{eqn:sig}
    \end{equation}
    
    Equation \eqref{eqn:sig} has two salient differences to \textit{IG} from Equation \eqref{eqn:ig}.
    First, the explanation point $z_i$ in the inner product $(x_i- z_i)$ is part of the integral, whereas in \textit{IG}, it is outside of it.
    Second, the integration set $S$ is not a path from $x$ to some point $z$ as it was in \textit{IG}.
    
    Similarly, for the reason of efficient computation, the integral can also be approximated using a discrete summation over $R$ multiple roots $z^{(r)}$:
    \begin{equation}
    SmoothTaylor_i(x) \approx \frac{1}{R} \sum_{r=1}^{R} (x_i - z^{(r)}_i) \frac{\partial f(z^{(r)})}{\partial x_i}, z^{(r)} \sim S 
    \label{eqn:sig-approx}
    \end{equation}
    
    Equation \eqref{eqn:sig-approx} is derived from the averaged Taylor's theorem approximation in Equation \eqref{eqn:taylor-series-average} by choosing a set of roots such that the model output score difference between each root $z^{(r)} \in S$ and input $x$ is almost close to zero (i.e. $\forall{r}: f(x) - f(z^{(r)}) \approx 0$).
    As a result, the inner summation term $f(z^{(r)})$ is canceled out with $f(x)$, and the remaining terms can be explained as the sum of the smooth integrated gradients across all dimensions.
    Note that this loosely satisfies the \textit{completeness} axiom just like the \textit{IG} method.
    It also fulfils the \textit{implementation invariance} property.
    
    The next issue is to decide on a suitable method to generate the roots $z^{(r)}$.
    If one is interested in classification or segmentation as pixel-wise classification, then one would want to choose the set $S$ to be a set of points where the prediction output class switches.
    However searching these points on the training dataset might result in roots which are too far away from the input $x$ to be explained, which will impact the quality of the Taylor approximation.
    One alternative is to seek for a random set of points sufficiently close to $x$, so that the quality of the Taylor approximation is acceptable, and also sufficiently far away, so that the noise from the gradient shattering effect in deep networks \cite{Balduzzi2017} can be canceled out by averaging over many $z$ from many different linearity regions.
    A simple approach, inspired by \textit{SmoothGrad}, is to add a random variable $\epsilon$ to input $x$, where $\epsilon$ can be drawn from a Gaussian distribution with standard deviation $\sigma$ being the noise scaling factor:
    \begin{equation}
    \label{eqn:roots-generation}
    z^{(r)} = x + \epsilon, \mathrm{where}\ \epsilon \sim \mathcal{N}(0,\sigma^2)
    \end{equation}
    
    The choice of the $\sigma$ value should be carefully selected, and it is further discussed in Section \ref{sec:adapt-noise}. This follows the principle of choosing $z^{(r)}$ to be close to $x$ and also sufficiently far away, so that the need for a good Taylor approximation and averaging effect of the noise in the gradients can be balanced.\\
    
    \noindent\textbf{Theorem:} If the roots in \textit{SmoothTaylor} are chosen as per Equation \eqref{eqn:roots-generation}, then the discrete version of \textit{SmoothTaylor} as given in Equation \eqref{eqn:sig-approx} is a special case of \textit{SmoothGrad} with $\mathcal{M} = \nabla f(x+\epsilon) \cdot \epsilon  $.\\
    
    This theorem does not hold for other choices of the set $S$ in Equation \eqref{eqn:sig-approx}, thus \textit{SmoothTaylor} defines an algorithm class of its own.
    
    \textit{SmoothTaylor} offers an alternative formulation to \textit{IG}, where the selection of a fixed baseline is not required. The above theorem establishes \textit{SmoothTaylor} with a choice of roots as in Equation \eqref{eqn:roots-generation} as a theoretical bridging concept between \textit{IG} and \textit{SmoothGrad}.
    
    \section{Experiments}\label{sec:experiment}
    
    We apply \textit{SmoothTaylor} and \textit{IG} \cite{Sundararajan2017} attribution techniques, and compare their results.
    We choose to analyze them on the image classification task.
    The goal is to compare the quality of the attribution maps computed by these two methods.
    To encourage reproducibility, we publicly release our source code\footnote{\url{https://github.com/garygsw/smooth-taylor}}.
    Here, we describe our experiment setup and evaluation metrics.
    
    \subsection{Setup}
    
    We use the first 1000 images from the ILSVRC2012 ImageNet object recognition dataset \cite{Russakovsky15} validation subset as the scope of our experiment.
    It is a 1000 multi-class image classification task, with each image preprocessed to be the size of $224 \times 224$ pixels.
    We choose two deep neural image classifier models, DenseNet121 \cite{Huang2017} and ResNet152 \cite{He2015}, that are both pretrained on the ImageNet dataset to apply the attribution methods.
    We compute the attributions with respect to the function of the predicted class for each input image regardless of the ground truth label.
    Therefore, the attribution process is entirely unsupervised.
    
    \begin{figure*}[!t]
        \begin{minipage}{.58\linewidth}
            \centering
            \includegraphics[width=.9\linewidth]{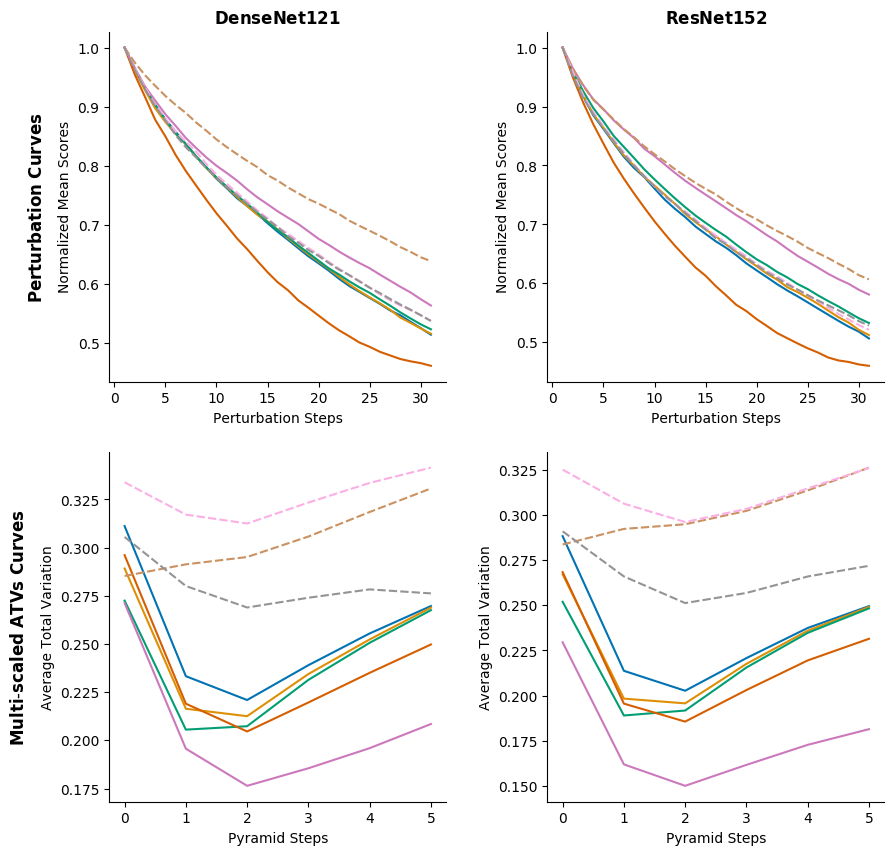}
        \end{minipage}
        \begin{minipage}{.3\linewidth}
            \centering
            \includegraphics[width=.9\linewidth]{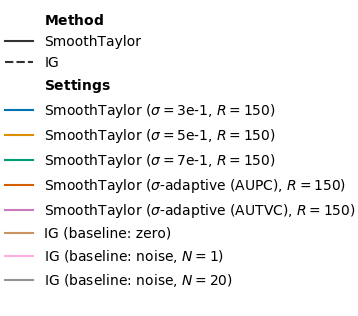}
            \includegraphics[width=.9\linewidth]{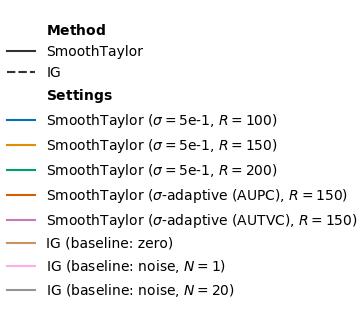}
        \end{minipage}
        \caption{Evaluation metrics curves; the lower the curve the better. \underline{Right}: Legends. \underline{Top row}: Perturbations curves. \underline{Bottom row}: Multi-scaled TV curves. \underline{Left column}: Based on DenseNet121. \underline{Right column}: Based on ResNet152.}
        \label{fig:curves}
    \end{figure*}
    
    \subsection{Hyperparameters}
    
    For the \textit{SmoothTaylor} method, we vary the parameter values for the number of roots $R$ to be 100, 150, and 200, and the noise scaling factor $\sigma$ to be \num{3e-1}, \num{5e-1}, and \num{7e-1}.
    The magnitudes of the noise scaling factor are decided to be roughly in the range of the average values of the inputs after normalization.
    For \textit{IG}, we choose total steps $M$ to be 50, and vary the type of baselines used.
    We use the zero (black image) baseline, and random uniform noise baselines with different samples sizes $N$ to be 1, 5, 10, and 20.
    
    \subsection{Evaluation Metrics}\label{sec:eval}
    
    Sundararajan et al. \cite{Sundararajan2017} argued against empirical methods for evaluating attribution methods, and thus decided to rely on an axiomatic approach to determine the quality of an attribution method.
    However, axiom sets might be incomplete, and for a data-driven science, a quantitative evaluation is often aligned with the goals. 
    Furthermore, there are limitations to qualitative evaluation of attribution maps due to biases in human intuition towards simplicity whereas deep neural models which might be over-parametrized and thus of high complexity.
    Therefore, in this paper, we use the following two quantitative metrics:
    
    \subsubsection{Perturbation Approach}
    
    One such metric suggested by Samek et al. \cite{Samek2017} relies on selecting the top salient regions of pixels in the input image by attribution and successively replacing them with random noise (also known as pixel perturbation), and then measuring the drop in model output scores.
    A higher score drop signifies a more sensitive attribution method, since the attributions are able to better identify the salient parts of the input that explain the model's output.
    
    We describe our pixel perturbation evaluation procedure formally as follows.
    First, we use a sliding local window of kernel size $k \times k$ in the input image space to find an ordered sequence $\mathcal{O} = (r_1, r_2, ..., r_L)$ that contains the top-$L$ most salient non-overlapping regions.
    The sorting of the regions is based on the average absolute attribution values of the pixels' location within each kernel window, from the highest to lowest (most relevant first).
    A high average absolute attribution value in a region $r_l$ denotes a high presence of evidence that supports the model's prediction.
    
    Second, we follow the sequence of ordered regions in $\mathcal{O}$ to apply the perturbations on.
    Let $g(x,r)$ be a function which performs the perturbation on some input image $x$ at region $r$, where information in that region is removed by the replacement of the value of its pixels with random values drawn from a uniform distribution across the valid input value range.
    The function $g$ is then successively applied starting with the original input image $x^{(0)} = x$.
    The input image for the next step $x^{(l)}$ is iteratively updated after perturbation at step $l$ for $L$ times:
    \begin{equation}
    \forall{\;1 \leq l \leq L}:\ x^{(l)} = g(x^{(l-1)}, r_l)
    \label{eqn:perturbation-process}
    \end{equation}
    At each step $l$, we consider $P$ number of different random perturbation samples and compute the mean score $\bar{y}^{(l)}$: 
    \begin{equation}
    \bar{y}^{(l)} = \frac{1}{P} \sum_{p=1}^{P} f(x^{(l-1)^{(p)}})
    \label{eqn:perturbation-mean}
    \end{equation}
    
    The perturbation with the median output score is selected as the actual perturbation to update.
    To quantitatively measure the strength of an attribution method, we look at how much these mean output scores drop with steps $l$.
    That can be quantified by taking the area under the perturbation curve (AUPC) (see Figure \ref{fig:curves} (top)) after normalizing each mean score $\bar{y}^{(l)}$ at each step $l$ with the original score $f(x)$, and averaged over all images in the dataset.
    Throughout our experiments, we use kernel size $k=15$, number of perturbations $L=30$, and perturbation sample size $P=50$.
    
    \subsubsection{Average Total Variation}
    
    We use average total variation (ATV) as the second evaluation metric to measure the smoothness or the total amount of noise of each pixel with its local neighbors.
    We consider a saliency map $\mathcal{S}$ as vector of size $h \times w$ to represent every pixel.
    Taking only absolute values, a min-max normalization (with values above 99\textsuperscript{th} percentile clipped off) is applied on an attribution map to construct a saliency map.
    The ATV of $\mathcal{S}$ is computed as follows:
    \begin{equation}
    ATV(\mathcal{S}) = \frac{1}{h \times w} \sum_{i,j \in \mathcal{N}} \Vert \mathcal{S}_i - \mathcal{S}_j \Vert_p
    \label{eqn:tv}
    \end{equation}
    Here, $\mathcal{N}$ defines the set of pixel neighbourhoods (adjacent horizontal and vertical pixels) and $\Vert\cdot\Vert$ is the $\ell_p$ norm.
    We use the established $\ell_1$-norm in our experiments.
    
    In addition, we construct Gaussian pyramids \cite{Burt1983} on the saliency maps by repeatedly scaling their dimensions down by 1.5 and applying a Gaussian smoothing filter to remove information.
    This process is repeated for each saliency map until the size of the map is smaller than $30 \times 30$ pixels.
    We then compute the ATV of the scaled and blurred saliency maps at each step -- we call them multi-scaled ATVs.
    Subsequently, after averaged over all images, we take the area under the multi-scaled ATVs curve (AUTVC) (see Figure \ref{fig:curves} (bottom)) as the measure quantity to evaluate the quality of an attribution method.
    
    \begin{table}[t!]
        \centering
        \caption{Area under the curves results. \hspace{\textwidth}Note: Lower AUPC and AUTVC is better.}
        \begin{tabularx}{240pt}{c*{6}{>{\centering\arraybackslash}X}}
            \toprule
            \multicolumn{2}{c}{\multirow{2}{*}{\textbf{Attribution Method}}} & \multicolumn{4}{c}{\textbf{Image Classifier Model}} \\
            \cmidrule{3-6}
            & & \multicolumn{2}{c}{DenseNet121} & \multicolumn{2}{c}{ResNet152}\\
            \midrule
            \multicolumn{2}{c}{\textbf{IG}} & & & &\\
            baseline & $N$ & AUPC & AUTVC & AUPC & AUTVC\\
            \midrule
            zero & - & 23.63 & 1.52 & 22.87 & 1.51\\
            \cmidrule{1-6}
            \multirow{4}{*}{noise}
            & $1$ & 21.51 & 1.62 & 21.05 & 1.54\\
            & $5$ & 21.54 & 1.52 & 20.99 & 1.43\\
            & $10$ & 21.46 & 1.45 & \textbf{21.02} & 1.37\\
            & $20$ & \textbf{21.43} & \textbf{1.39} & \textbf{21.02} & \textbf{1.32}\\
            \midrule
            \multicolumn{2}{c}{\textbf{SmoothTaylor}}
            & \multicolumn{2}{c}{DenseNet121} & \multicolumn{2}{c}{ResNet152}\\
            \cmidrule{3-6}
            $\sigma$ & $R$ & AUPC & AUTVC & AUPC & AUTVC\\
            \midrule
            \multirow{3}{*}{\num{3e-1}}
            & $100$ & 21.24 & 1.28 & 20.83 & 1.20\\
            & $150$ & 21.19 & 1.24 & 20.79 & 1.16\\
            & $200$ & \textbf{21.13} & 1.22 & \textbf{20.78} & 1.14\\
            \cmidrule{1-6}
            \multirow{3}{*}{\num{5e-1}}
            & $100$ & 21.25 & 1.23 & 21.00 & 1.14\\
            & $150$ & 21.20 & 1.19 & 20.95 & 1.10\\
            & $200$ & \textbf{21.13} & 1.16 & 20.86 & 1.07\\
            \cmidrule{1-6}
            \multirow{3}{*}{\num{7e-1}}
            & $100$ & 21.39 & 1.20 & 21.37 & 1.08\\
            & $150$ & 21.30 & 1.15 & 21.32 & 1.04\\
            & $200$ & 21.30 & \textbf{1.12} & 21.14 & \textbf{1.01}\\
            \bottomrule
        \end{tabularx}
        \label{tab:auc-results}
    \end{table}
    
    \begin{table}[t!]
        \centering
        \caption{Area under the curves results for \textit{SmoothTaylor} with extreme hyperparameter values.\hspace{\textwidth}Note: Lower AUPC and AUTVC is better.}
        \begin{tabularx}{240pt}{c*{6}{>{\centering\arraybackslash}X}}
            \toprule
            \multicolumn{2}{c}{\textbf{SmoothTaylor}} & \multicolumn{4}{c}{\textbf{Image Classifier Model}} \\
            \cmidrule{3-6}
            \multicolumn{2}{c}{\textbf{Hyperparameters}} & \multicolumn{2}{c}{DenseNet121} & \multicolumn{2}{c}{ResNet152}\\
            \midrule
            $\sigma$ & $R$ & AUPC & AUTVC & AUPC & AUTVC\\
            \midrule
            \num{5e-1} & $10$ & 21.74 & 1.55 & 21.43 & 1.43\\
            \cmidrule{1-6}
            \num{1e-4} & $100$ & 23.45 & 1.79 & 23.00 & 1.55\\
            \num{1e-3} & $100$ & 23.60 & 1.53 & 23.14 & 1.48\\
            \num{1e-2} & $100$ & 23.90 & 1.57 & 23.46 & 1.23\\
            \num{1e-1} & $100$ & 22.03 & 1.43 & \textbf{21.44} & 1.22\\
            \num{1e+0} & $100$ & \textbf{21.88} & \textbf{1.17} & 22.16 & \textbf{1.04}\\
            \num{2E+0} & $100$ & 23.54 & 1.19 & 24.48 & 1.27\\
            \bottomrule
        \end{tabularx}
        \label{tab:extreme-results}
    \end{table}

    \begin{table}[t!]
        \centering
        \caption{Area under the curves results with Adaptive Noising. \hspace{\textwidth}Note: Lower AUPC and AUTVC is better.}
        \begin{tabularx}{240pt}{c*{6}{>{\centering\arraybackslash}X}}
           \toprule
            \multicolumn{2}{c}{\textbf{SmoothTaylor}} & \multicolumn{4}{c}{\textbf{Image Classifier Model}} \\
            \cmidrule{3-6}
            \multicolumn{2}{c}{\textbf{Hyperparameters}} & \multicolumn{2}{c}{DenseNet121} & \multicolumn{2}{c}{ResNet152}\\
            \midrule
            $\sigma$ & $R$ & AUPC & AUTVC & AUPC & AUTVC\\
            \midrule
            Adaptive-AUPC & $150$ & \textbf{19.55} & 1.14 & \textbf{19.30} & 1.05\\
            Adaptive-AUTVC & $150$ & 22.14 & \textbf{0.99} & 22.52 & \textbf{0.85}\\
            \bottomrule
        \end{tabularx}
        \label{tab:auc-results}
    \end{table}

    \subsection{Results}
    
    We compute the attribution maps using a few different attribution methods based on two pretrained image classifiers on the ImageNet dataset.
    Examples of these attribution maps are visualized as saliency maps in Figure \ref{fig:method-comparison}.
    
    Qualitatively, we can observe that \textit{SmoothTaylor} produces visually sharper saliency maps as compared to \textit{IG}.
    In addition, they are better at highlighting distinctive regions that explain the model's prediction.
    While it is not the best method that produces the least noise or the most sensitivity (see saliency maps produced by Layer-wise Relevance Propagation \cite{Binder2016}), \textit{SmoothTaylor} offers ease of implementation and fulfils the two current fundamental axioms of an attribution method.
    
    Next, we discuss the results using quantitative evaluation measures.
    A summary of the experimental results is shown in Table \ref{tab:auc-results} with the AUPC and AUTVC values for each experiment run.
    The Simpson's rule is used to compute the area under the curves.
    We analyze the results based on two objectives -- sensitivity and noise level, and also compare the results based on two different classifier models.
    
    \subsubsection{Sensitivity}
    
    As observed in Figure \ref{fig:curves} (top), when compared to \textit{IG}, the attribution maps of \textit{SmoothTaylor} are able to cause a larger classification score drop as perturbation step increases.
    Expectedly, the AUPC values for \textit{SmoothTaylor} are also lower, showing that \textit{SmoothTaylor} is more sensitive to relevant explanations points in the input space than \textit{IG}.
    The averaged \textit{IG} with noise baselines are shown to have large improvements; almost close to the performance of \textit{SmoothTaylor} at our chosen hyperparameters, though still a little worse.
    Their improvements also produce diminishing marginal returns as $N$ increases beyond more than 5.
    On closer inspection with Table \ref{tab:auc-results}, it shows that our choice for $\sigma$ values did not produce any significant effect on the AUPC values, which is worth investigating further in Section \ref{sec:extreme}.
    However, the AUPC values clearly decrease as $R$ increases.
    This is expected as the ``smoothing" effect is greater when we draw more roots, resulting in a statistically better representation of $z$ which improves the power of the Taylor approximation.
    
    \subsubsection{Noise level}
    
    The \textit{SmoothTaylor} method clearly generates attribution maps that are much less noisy than \textit{IG}.
    As seen in multi-scaled ATV curves in Figure \ref{fig:curves} (bottom), all the curves for \textit{SmoothTaylor} are lower that the curves for \textit{IG}.
    We also compare the effect of $\sigma$ and $R$ on the noisiness of the attribution maps of \textit{SmoothTaylor}.
    First, the AUTVC values decrease as $R$ increases.
    This is also expected due to the increase ``smoothing" effect.
    Second, the AUTVC values seem to increase as $\sigma$ increases.
    However, we believe that this relationship is not monotonically true, as the selection of our $\sigma$ values may be too low across all images in the dataset.
    We discuss this further in Section \ref{sec:extreme}.
    
    \subsubsection{DenseNet121 vs. ResNet152}
    
    The sensitivity improvements in the perturbation curves by \textit{SmoothTaylor} over \textit{IG} is noticeably lesser for ResNet152 as compared to DenseNet121.
    One hypothesis is that the gradients from ResNet152 are less noisy to begin with, since residual networks are shown to have reduced shattering gradients effect.
    Thus, with more reliable gradients to explain for the model's prediction, the effectiveness of smoothing is also reduced.
    
    \begin{figure*}[!t]
        \centering
        \begin{minipage}{.57\linewidth}
            \centering
            \includegraphics[width=.9\linewidth]{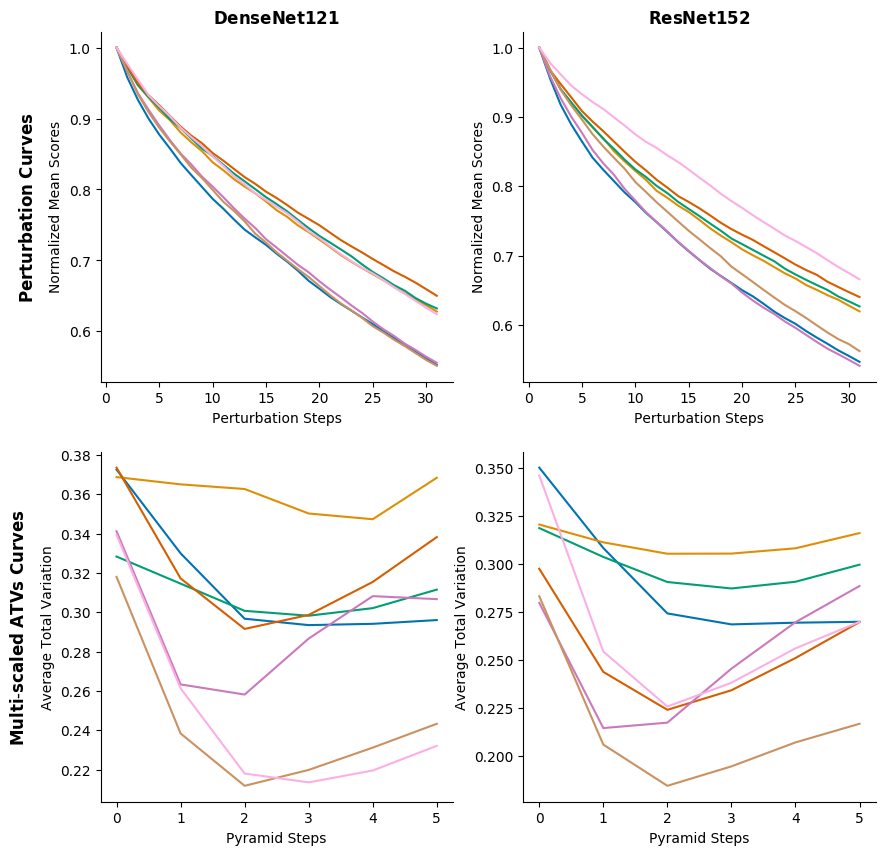}
        \end{minipage}
        \begin{minipage}{.3\linewidth}
            \centering
            \includegraphics[width=.9\linewidth]{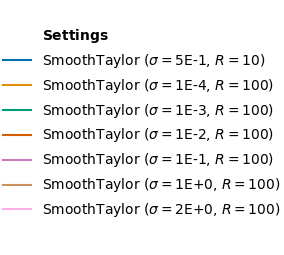}
        \end{minipage}
        \caption{Evaluation metrics curves for the study of the impact of varying the noise hyperparameter; the lower the curve the better. \underline{Top row}: Perturbation curves. \underline{Bottom row}: Multi-scaled TV curves. \underline{Left column}: Based on DenseNet121. \underline{Right column}: Based on ResNet152.}
        \label{fig:extreme-curves}
    \end{figure*}
    
    \subsection{Noise Hyperparameter Sensitivity Analysis}\label{sec:extreme}
    
    We choose a range of $\sigma$ values as high as \num{2e+0} and as low as \num{1e-4}, while fixing $R$ to be 100.
    The effects of different values of the noise scale parameter for \textit{SmoothTaylor} are displayed in Figure \ref{fig:extreme-curves}, and its results are summarized in Table \ref{tab:extreme-results}.
    
    
    
    We can observe that for too small noise choices such as \num{1e-4} or \num{1e-3}, the AUPC sensitivity is lower than for choices in the order of \num{1e-1}.
    This can be explained from the effect of gradient shattering in deep networks: when the gradient has a large component resembling white noise, as observed in \cite{Balduzzi2017}, then using averages is a statistically reasonable attempt to remove the white noise component.
    Rectified Linear Units (ReLu) networks consist of zones with locally linear predictions -- see Figure 3 in \cite{Novak2018} for a clear illustration of this effect.
    
    The gradient is constant within each such zone.
    Above averaging requires to sample the gradient at many different local linearity zones around the sample of interest $x$.
    In particular averaging requires $z_i$ to be outside of the linearity zone in which $x$ is in.
    This explains why a very small amount of noise will not result in an effective averaging of white noise, as most of the samples $z_i$ would just stay in the local linearity zone of $x$ and fail to sample different gradient values.
    
    The size of the local linearity zone is sample-dependent \cite{Novak2018}.
    This observation supports the claim that the noise scale $\sigma$ needs to be carefully calibrated within a certain range (i.e. it cannot be too small or too big) for every individual sample $x$ in order for the attribution maps of \textit{SmoothTaylor} to be of high quality.
    Therefore, based on this observation, we go further and propose an adaptive improvement to \textit{SmoothTaylor} in the next section.
    
    \section{Adaptive Noising}\label{sec:adapt-noise}
    
    Ideally, the value of noise scale $\sigma$ should depend on each individual input, and not generally fixed to all inputs.
    Thus, we propose an adaptive noising technique to search for an optimal noise scale value for each input, so as to optimize the \textit{SmoothTaylor} method.
    
    We adopt an iterative heuristic line search approach to design our algorithm.
    The goal is to find an optimal value for $\sigma$ such that the attribution maps can be the most sensitive or least noise (quantified by AUPC or AUTVC respectively).
    As such, while fixing $R$, we search for $\sigma^*$ for each input such that the AUPC or AUTVC of its attribution map is minimized.
    We describe our algorithm in Algorithm \ref{alg:adaptive-noising}.
    
    \begin{algorithm}[t!]
        \SetAlgoLined
        \scriptsize
        \SetKwInOut{Input}{Input}
        \SetKwInOut{Output}{Output}
        \SetKwInOut{Parameters}{Parameters}
        \SetKwFunction{ComputeAUC}{ComputeAUC}
        \Parameters{Max. iterations $i_{max}$, learning rate $\alpha$, learning decay $\gamma$, max. stop count $s_{max}$}
        \Input{Input $x$, root size $R$, model $f$}
        \Output{Optimal $\sigma^*$ value}
        \Begin{
            $\sigma \gets \frac{1}{N} \sum |x|$\;
            $\mathrm{AUC} \gets $ \ComputeAUC{$x, R, f, \sigma$}\; 
            $i \gets 1$; $s \gets 0$; $\sigma^* \gets \sigma$; $\mathrm{AUC}^* \gets \mathrm{AUC}$\;
            \BlankLine
            \While{$i \leq i_{max}$}{
                $\mathrm{AUC}_s \gets $ \ComputeAUC{$x, R, f, |\sigma + \alpha|$}\;
                \eIf{$\mathrm{AUC}_s > \mathrm{AUC}$}{
                    $\sigma \gets |\sigma - \alpha|$\;
                    $\mathrm{AUC}_s \gets $ \ComputeAUC{$x, R, f, \sigma$}\;
                }{
                    $\sigma \gets |\sigma + \alpha|$\;
                }
                \BlankLine
                \eIf{$\mathrm{AUC}_s > \mathrm{AUC}$}{
                    \eIf{$s \leq s_{max}$}{
                        $\alpha \gets \alpha * \gamma$; $s \gets s + 1$\;
                    }{
                        \textbf{break}
                    }
                }{
                    $s \gets 0$\;
                    \If{$\mathrm{AUC}_s < \mathrm{AUC}^*$}{
                        $\mathrm{AUC}^* \gets \mathrm{AUC}_s$; $\sigma^* \gets \sigma$\;
                    }
                }
                $\mathrm{AUC} \gets \mathrm{AUC}_s$; $i \gets i + 1$\;
            }
        }
        \caption{Adaptive Noising}
        \label{alg:adaptive-noising}
    \end{algorithm}
    
    In our proposed iterative optimization procedure, we search for $\sigma^*$ within maximum iterations of $i_{max}$.
    We include an early stopping mechanism with maximum stop count $s_{max}$.
    At each iteration, $\sigma$ is updated with learning rate $\alpha$ which direction depends on a line search.
    The learning rate is reduced by a factor learning decay $\gamma < 1$ whenever the current iteration's AUC is greater than the previous one.
    In our experiment, we use $R=150$ and set maximum iterations $i_{max}=20$, maximum stop count $s_{max}=3$, learning rate $\alpha=0.1$, learning decay $\gamma=0.9$, and use the same setup from the AUC computation in our earlier experiments.
    
    We report the results from using adaptive noising in Table \ref{tab:auc-results} and compare with the results from previous experiment runs.
    With adaptive noising, we are able to obtain the best AUPC or AUTVC values among all runs.
    However, it is to be noted that computing AUPC is computationally expensive and slow while computing AUTVC is much faster.
    The results conclusively show that \textit{SmoothTaylor} with adaptive noising is preferable over constant noise injection.
    
    \section{Conclusion}\label{sec:conclusion}
    
    Explaining for all deep neural model decisions is a huge challenge given the vast taxonomy of model types and scope of problems.
    Thus it is crucial to find a simple attribution method that is easily applied to various model architectures so as to encourage widespread usage.
    In this paper, we bridge \textit{IG} and \textit{SmoothGrad} and proposed \textit{SmoothTaylor} from the Taylor's theorem perspective.
    In our experiments, we also introduce multi-scaled average total variation as a new measure for noisiness of saliency maps.
    We further proposed adaptive noising as a hyperparameter tuning technique to optimize our proposed method's performance.
    From the experimental results, \textit{SmoothTaylor} is able to produce attribution maps that are more relevance-sensitive and with much less noise as compared to \textit{IG}.
    
    \section*{Acknowledgment}
    
    The research is supported by the National Research Foundation, Prime Minister's Office, Singapore, under its CREATE programme, Singapore-MIT Alliance for Research and Technology (SMART) Future Urban Mobility (FM) IRG.

    \bibliographystyle{IEEEtran}
    \bibliography{references}
    
\end{document}